\documentclass[10pt,twocolumn,letterpaper]{article}

\usepackage[pagenumbers]{cvpr}
\definecolor{cvprblue}{rgb}{0.21,0.49,0.74}
\usepackage[pagebackref,breaklinks,colorlinks,allcolors=cvprblue]{hyperref}
\usepackage{algorithm}
\usepackage{algorithmic}
\usepackage{multirow}
\usepackage{amsmath}
\usepackage{mathtools}
\usepackage{amsthm}
\usepackage{booktabs}
\usepackage[table]{xcolor}
\usepackage{colortbl}
\usepackage{pgfplots}
\pgfplotsset{compat=1.18}
\usepgfplotslibrary{groupplots}

\def\paperID{*****} 
\def\confName{CVPR}
\def\confYear{2026}

\def\eg{\emph{e.g.}}

\def\ie{\emph{i.e.}}
\definecolor{gen}{RGB}{100,100,100}
\definecolor{hl}{RGB}{240,240,240}

\title{Detecting Diffusion-generated Images via \texttt{D}ynamic \texttt{A}ssembly \texttt{F}orests}

\author{
Mengxin Fu \quad Yuezun Li\thanks{Corresponds to Yuezun Li (\url{liyuezun@ouc.edu.cn})} \\
School of Computer Science and Technology, Ocean University of China
}

\begin{document}
\maketitle
\begin{abstract}
Diffusion models are known for generating high-quality images, causing serious security concerns. To combat this, most efforts rely on deep neural networks (\eg, CNNs and Transformers), while largely overlooking the potential of traditional machine learning models. In this paper, we freshly investigate such alternatives and proposes a novel \texttt{\textbf{D}}ynamic \texttt{\textbf{A}}ssembly \texttt{\textbf{F}}orest model (\texttt{\textbf{DAF}}) to detect diffusion-generated images. Built upon the deep forest paradigm, \texttt{\textbf{DAF}} addresses the inherent limitations in feature learning and scalable training, making it an effective diffusion-generated image detector. Compared to existing DNN-based methods, \texttt{\textbf{DAF}} has significantly fewer parameters, much lower computational cost, and can be deployed \textit{without} GPUs, while achieving competitive performance under standard evaluation protocols. These results highlight the strong potential of the proposed method as a practical substitute for heavyweight DNN models in resource-constrained scenarios. Our code and models are available at \url{https://github.com/OUC-VAS/DAF}.
\end{abstract}    
\section{Introduction}
\label{sec:intro}

Diffusion models underpin most recent mainstream generative techniques and are capable of producing high-quality images under diverse conditions~\cite{liu2022pseudo, li2023gligen, yang2024mastering}. While these models enable beneficial applications such as virtual reality and cinematic visual effects, their misuse can lead to misinformation and fabricated content, substantially undermining human trust in digital media~\cite{cozzolino2024zero,Chen2024ASS,Wang_2025_CVPR}. To counter this issue, many efforts have been proposed to detect diffusion-generated images~\cite{Ricker_2024_CVPR,Cazenavette_2024_CVPR,Chu_2025}. Given the high realism of generated images, mainstream methods commonly rely on deep neural networks (\eg, CNNs and Transformers) to capture underlying generative traces and have shown favorable performance on standard datasets~\cite{Wang_2023_ICCV, 5206848, yu15lsun}.

\begin{figure*}
    \centering
    \includegraphics[width=\linewidth]{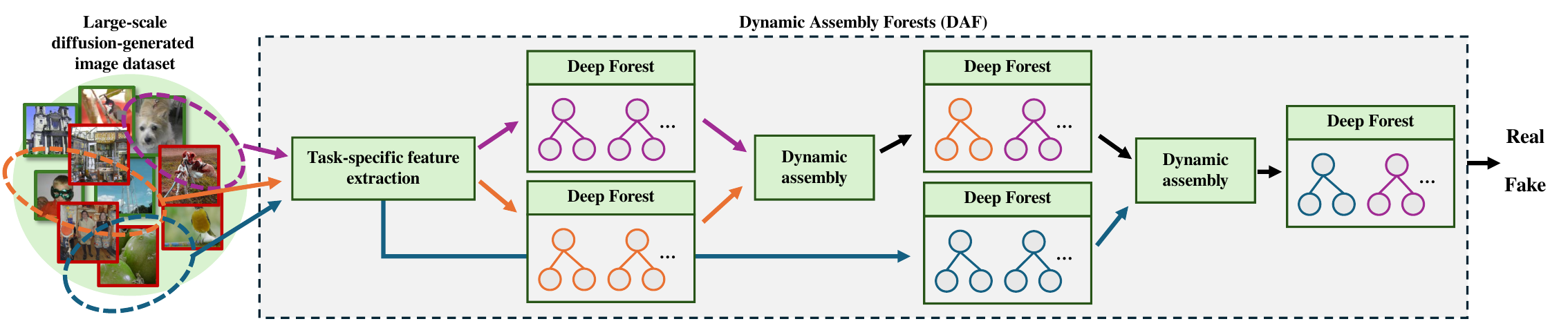}
    \caption{\small The overview of our proposed \texttt{\textbf{DAF}} model.}
    \label{fig:overview}
\end{figure*}

In this paper, we investigate a previously underexplored direction: detecting diffusion-generated images with traditional machine learning rather than DNNs. Generally, machine learning models are considered less effective than DNNs due to their limited model capacity, fewer parameters, and simpler structures, and therefore have attracted little attention in this task. Nevertheless, we rethink this assumption and propose a novel forest-based model termed \texttt{\textbf{D}}ynamic \texttt{\textbf{A}}ssembly \texttt{\textbf{F}}orest model (\texttt{\textbf{DAF}}). Our method is built upon the deep forest paradigm~\cite{ijcai2017p497, lu2024forensicsforest} and introduces two significant improvements that enable effective detection of diffusion-generated images (see Fig.~\ref{fig:overview}).

First, we introduce a dynamic assembly strategy that constructs a forest model
in a batch-wise training manner analogous to the training procedure of DNNs. Note that in the original deep forest paradigm, all training samples must be loaded into memory simultaneously, which becomes impractical for large-scale datasets. To overcome this limitation, our strategy randomly samples a small subset of training images via bootstrapping and trains an individual forest model on each subset. This process is repeated multiple times to obtain a small collection of forest models. These independently trained forests are then dynamically assembled into a unified model by selecting and ensembling the best-performing components from each forest model. Next, we sample data again and repeat the above process until a predefined maximum time is reached or no significant improvement is obtained. This dynamic assembly strategy substantially reduces the one-time memory overhead and therefore enables the adoption of a more advanced feature extraction strategy.

Second, we describe a task-specific feature extractor tailored to generative traces. In the deep forest paradigm, model complexity grows with the dimensionality of the input features, which necessitates a careful trade-off between performance and model scale. The work~\cite{lu2024forensicsforest} achieves this by highly compressing the input image with simple color and frequency histograms. While effective in reducing feature dimensionality, such a strategy mainly captures global information and overlooks fine-grained details, thus hindering the ability to capture subtle generative traces. To address this, we describe a patch-based strategy that jointly considers both the global and fine-grained information. Specifically, the input image is partitioned into multiple small patches, from which spatial and frequency-related features are extracted individually. Then, multi-scale features are considered by fusing patches at different scales. Thanks to the proposed Dynamic Assembly Strategy, the advanced feature extraction can be deployed without being limited by memory overhead.

Compared to existing DNN-based methods, \texttt{\textbf{DAF}} requires significantly fewer parameters, incurs substantially lower computational cost, and can be deployed \textit{without} GPUs. Extensive experiments demonstrate that our method achieves competitive -- and in some cases superior -- performance compared to DNN counterparts. These findings provide new insights into forensics-oriented machine learning research and highlight the strong potential of \texttt{\textbf{DAF}} as a practical alternative for heavyweight DNN models in resource-constrained scenarios. Our contribution can be summarized as follows:
\begin{itemize}
    \item Departing from the DNN-dominated paradigm in modern forensics, we demonstrate the feasibility of traditional machine learning algorithms for diffusion-generated image detection, encouraging further forensics-oriented machine learning research.
    \item We propose a novel forest-based model named Dynamic Assembly Forests (\texttt{\textbf{DAF}}), which enables forests to serve as effective detectors with two key improvements in forest construction and feature extraction. To the best of our knowledge, this is a fresh attempt to break the paradigm of static forest construction by introducing a dynamic assembly mechanism for diffusion-generated image detection. This design significantly expands the applicability of deep forest models and, in turn, unlocks the potential for more advanced feature extraction without being constrained by memory limitations. 
    \item Compared to existing DNN-based methods, our method employs significantly fewer parameters, incurs substantially lower computational cost, and can be deployed entirely on CPUs. Extensive experiments on standard datasets demonstrate its efficacy, highlighting its potential as a practical alternative to heavyweight DNN models in resource-constrained scenarios.
\end{itemize}

\begin{figure*}[!t]
    \centering
    \includegraphics[width=0.8\linewidth]{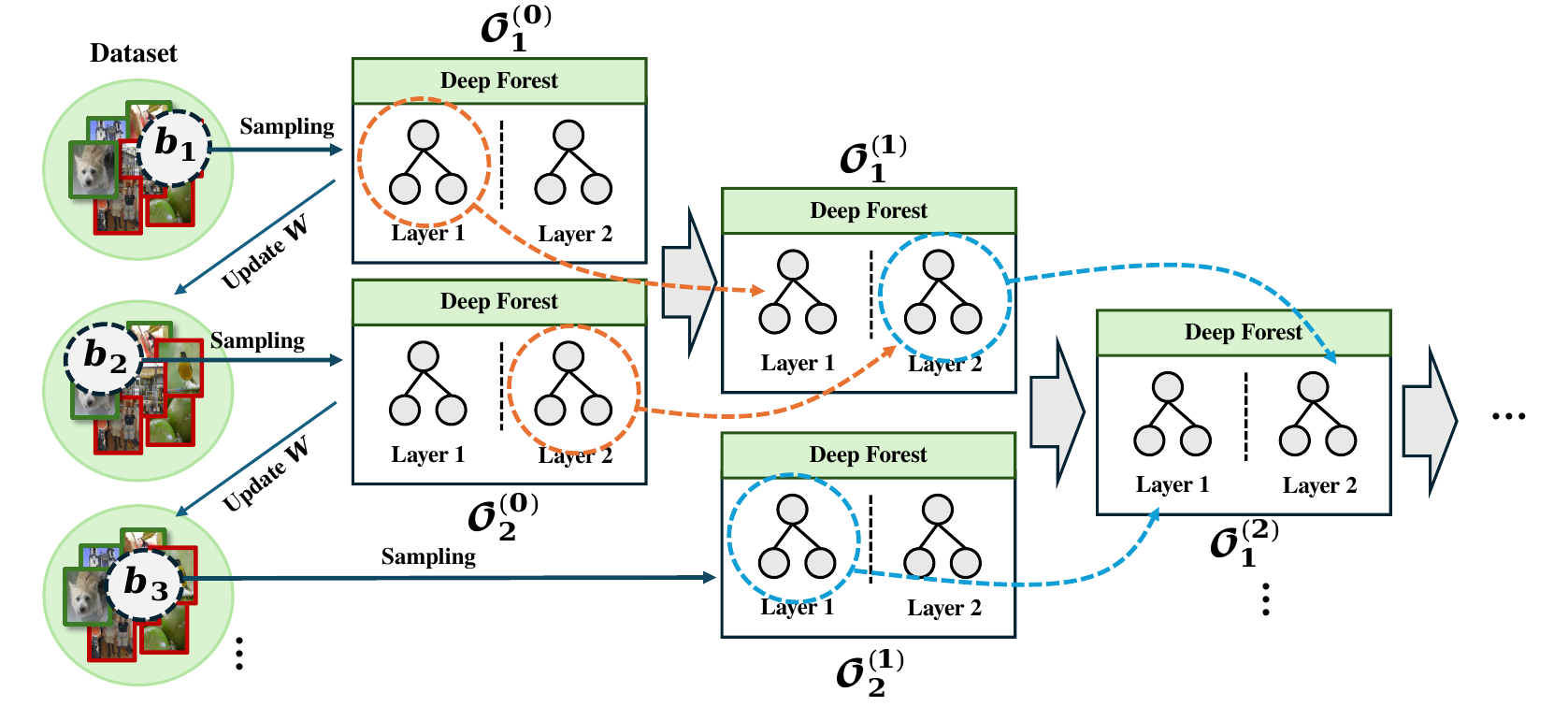}
    \vspace{-0.5cm}
    \caption{\small The detailed explanation of our proposed dynamic assembly strategy. }
    \label{fig:assembly}
\end{figure*}

\section{Related Works}
\label{sec:related}

\textbf{Diffusion-generated Image Detection:} 
Diffusion models represent the most recent and advanced class of generative techniques that could synthesize images with remarkably high realism~\cite{NEURIPS2020_4c5bcfec, Rombach_2022_CVPR, DBLP:conf/iclr/SongME21, NEURIPS2021_49ad23d1}. Their rapid progress has raised serious security and privacy concerns~\cite{10.1145/3746027.3755398,10.1145/3746027.3754567}. To cope with issue, many efforts have been proposed, \eg,~\cite{Wang_2023_ICCV,Ojha_2023_CVPR,10.1145/3746027.3755707,pmlr-v235-chen24ay,luo2024lare,huang2025diffusion,kang2025semantic,Chu_2025}. These methods typically design specific deep neural networks dedicated to exposing synthesis traces. We briefly review several representative methods below. For instance, CNNDet~\cite{wang2019cnngenerated} trains a universal CNN-fake detector on ProGAN that generalizes to many unseen generators. AEROBLADE~\cite{Ricker_2024_CVPR} detects diffusion-generated images by assessing autoencoder reconstruction error. FakeInversion~\cite{Cazenavette_2024_CVPR} detects synthetic images by inverting a fixed Stable Diffusion model and training a detector on the resulting inversion artifacts, enabling strong generalization to unseen text-to-image generators. DIRE~\cite{Wang_2023_ICCV} detects diffusion-generated images by DDIM inversion and reconstruction, and separates real from synthetic images based on the resulting reconstruction error. More recently, FIRE~\cite{Chu_2025} leverages the frequency decomposition of images by analyzing the variation in reconstruction errors before and after frequency filtering, further enhancing detection performance.

Despite their effectiveness, these methods mainly rely on DNNs, which require a heavy amount of computing resources. This complexity hinders their deployment in resource-constrained scenarios. Therefore, we investigate diffusion-generated image detection from a new perspective, exploring the use of traditional machine learning models as lightweight yet effective alternatives.

\textbf{Deep Forest and Related Models:} 
Deep forest~\cite{ijcai2017p497} was originally proposed for image classification tasks (\eg, MINIST~\cite{726791}). In contrast to conventional random forests, it adopts a cascade architecture composed of multiple forest layers, similar to DNNs. Each layer consists of several random and completely random forests, and the outputs of one layer are concatenated with the original features and passed to the subsequent layer for feature enhancement.
Despite its promise, deep forest has mainly been validated on low-dimensional tasks. Its application to high-dimensional tasks, such as high-quality generated image detection, remains highly limited for two main reasons: 1) it requires loading the entire training set into memory, which is impractical for large-scale datasets, and 2) it lacks effective input feature extraction strategies to capture generative traces.

ForensicsForest~\cite{lu2024forensicsforest} extends deep forest for GAN-generated face image detection. To obtain input features, it heavily compresses images using simple color and frequency histograms (It also use landmark for generated face images). While it shows effectiveness, it has only been evaluated on earlier StyleGAN architectures~\cite{Karras_2020_CVPR,8953766,NEURIPS2021_076ccd93}. Since diffusion models are more advanced, such a coarse strategy struggles to capture subtle traces in diffusion-generated images. Moreover, while this method alleviates memory consumption by training forests on data subsets and ensembling them in the end, it does not fundamentally address the concern, as the model ensemble itself also introduces a heavy memory cost. To this end, we propose \textbf{\texttt{DAF}}, which enables effective diffusion-generated image detection.
\section{Final copy}
\label{sec:proposed}

Existing forest-based models typically assume that all training samples are loaded into memory at once, which becomes impractical for diffusion-generated image detection tasks that require large-scale learning. 
Inspired by batch-wise learning in deep neural networks, we attempt to reformulate forest construction into a new batch-based training style, which can greatly increase the learning capacity under limited memory size. But a key challenge lies in the fact that, unlike deep neural networks whose parameters are differentiable and can be continuously updated across batches, forest models are inherently non-differentiable. As a result, enabling batch-wise updates for forest models is non-trivial and remains largely unexplored.
To address this challenge, we propose a dynamic assembly strategy that enables batch-wise training of forest models.

\smallskip
\noindent\textbf{Batches with Forest Construction.} 
We form a ``batch'' by randomly sampling a small subset of the entire training set and construct a deep forest model on this batch. Denote a batch as $b_i \subseteq D_{train}, |b_i| / |D_{train}| = p$, where $p$ denote the sampling ratio. Let $\mathcal{O}_i$ be the constructed forest model on $b_i$.
Since we could not follow the batch-by-batch serial update scheme in DNNs, we adopt ``bagging'' spirit, that first obtain multiple batches $\{b_i \}^v_{i=1}$ and then constructs different candidate forest models $\{\mathcal{O}_i \}^v_{i=1}$ accordingly, which are then jointly fused into a new forest model by dynamic assembly strategy.

\smallskip
\noindent\textbf{Forests Assembly.}
Our goal is to fuse all forest models $\{\mathcal{O}_i \}^v_{i=1}$ as a new forest model $\mathcal{O}'_v$, ensuring the model scale equivalent to a single forest model while achieving favorable performance. To this end, we select the most effective components from $\{\mathcal{O}_i \}^v_{i=1}$ and assemble them into a new forest model.

\noindent\textit{Question: How to select the most effective components? } 

Assume that each forest model in $\{\mathcal{O}_i \}^v_{i=1}$ consists of $l$ layers, where each layer contains $A$ random forests and $B$ completely random forests. To select components, we evaluate the performance of each forest model at the layer level. Specifically, we construct a validation subset independent of the batch data, denoted as $b_{val} \subseteq D_{train} \setminus \{b_i \}^v_{i=1}$.

In each corresponding layer of $\{\mathcal{O}_i \}^v_{i=1}$, we select in total $A$ random forests that achieve the highest accuracy on subset $b_{val}$ from all random forests, and similarly select $B$ completely random forests. These $A+B$ selected components form the structure of the corresponding layer in the assembled forest model. This procedure is applied to all layers, resulting in the final assembled model $\mathcal{O}^{(n)}_{i=1}$, where $n$ denotes the index of forest assembly. For example, $n=1$ indicates that the first forest assembly has been completed, whereas $n=0$ denotes no assembly is used. 
For clarity, we omit the superscript $(n)$ when first introducing $\{\mathcal{O}_i \}^v_{i=1}$.

Subsequently, we sample the next set of batches $\{b_i \}^{v}_{i=2}$ and repeat the forest construction process within each batch, producing another collection of forest models $\{\mathcal{O}^{(n)}_i \}^{v}_{i=2}$. The dynamic assembly strategy is then applied again to the set $\{ \mathcal{O}^{(n)}_1, \mathcal{O}^{(n)}_{2},..., \mathcal{O}^{(n)}_{v}\}$, yielding a new assembled model $\mathcal{O}^{(n+1)}_{i=1}$.

\begin{algorithm}[!t]
\caption{\small Dynamic Assembly Strategy}
\label{alg:DAF}
\begin{algorithmic}[1]
\STATE \textbf{Input:} Training set $D_{train}$, maximum assembly number $N$, forest number for assembly $v$, layer number $l$, random forest number $A$, completely random forest number $B$, initialized sample weights $W$, weight adjusting factor $\theta$, sampling ratio $p$
\STATE Initialize $n = 0$, $\mathcal{C} = \emptyset$
\WHILE{$n \leq N$}
\FOR{$i = 1:v$}
\IF{$\mathcal{C} \neq \emptyset$}
\STATE \textbf{Continue}
\ENDIF
\STATE $b_i \gets \mathrm{Random Sampling}(D_{train}, p, W)$ 
\STATE $\mathcal{O}^{(n)}_{i} \gets \mathrm{ForestConstruction}(b_i, l, A, B)$
\STATE $\mathcal{C} = \mathcal{C} \cup \mathcal{O}^{(n)}_{i}$
\STATE $b_{ws} \sim D_{train} \setminus b_i$
\STATE $W \gets \mathrm{DataWeightUpdate}(W, \theta, \mathcal{O}^{(n)}_{i}, b_{ws})$
\ENDFOR
\STATE $b_{val} \sim D_{train} \setminus \{b_i \}^v_{i=1}$
\STATE $\mathcal{O}^{(n+1)}_{i=1} \leftarrow \mathrm{ForestAssembly}(\mathcal{C},l,b_{val})$
\STATE $\mathcal{C} = \{\mathcal{O}^{(n+1)}_{i=1} \}$
\IF{EarlyStop($\mathcal{O}^{n+1}_{i=1}, \mathcal{O}^{n}_{i=1}$)}
\STATE \textbf{break}
\ENDIF
\STATE $n=n+1$
\ENDWHILE
\STATE \textbf{return} $\mathcal{O}^{n}_{i=1}$
\end{algorithmic}
\end{algorithm}

\smallskip
\noindent\textbf{Weighted Sampling.}
We aim for forests constructed in later batches to complement earlier ones, \ie, to better identify samples that were previously misclassified. To achieve this, we design a weighted sampling strategy that assigns adaptive weights to training samples prior to batch sampling, ensuring that each batch remains representative of the current learning state.

Initially, all the samples are assign equal weights, \ie, $W = \{w_i\}^{|D_{train}|}_{i=1}$, where $w_1 = \cdots = w_{|D_{train}|}$. After obtaining a forest model $\mathcal{O}_i$, we randomly sample a subset $b_{ws} \subseteq D_{train} \setminus b_i, |b_{ws}| \ll |D_{train} \setminus b_i|$ and and evaluate its prediction outcomes. For samples that are correctly classified, their weights are reduced, as they are relatively easy to detect. In contrast, the weights of misclassified samples are increased. Specifically, the weights are updated by multiplying or dividing by a factor $\theta > 1$, \ie, $w_i = w_i * (/) \theta$. This process is outlined in Fig.~\ref{fig:assembly}.

\smallskip
\noindent\textbf{Overall Algorithm.}
The overall procedure proceeds as follows. Firstly, a batch of data is sampled to train a deep forest model. The sample weights are then updated based on the performance of the trained model. After processing several batches, the dynamic assembly strategy is applied to fuse the individual forest models into a new model. This process is iterated multiple times until either a maximum number $N$ of assembly is reached or no significant performance improvement is observed (\ie, early stop by comparing $\mathcal{O}^{n+1}_{i=1}$ and $\mathcal{O}^{n}_{i=1}$). The complete workflow is summarized in Alg.~\ref{alg:DAF}.

\subsection{Task-specific Feature Extraction}
\label{featureextraction}
As generative traces are typically subtle, coarse feature extraction strategies, such as color and frequency histograms, are often ineffective. While fine-grained features can improve detection performance, they usually entail high feature dimensionality, leading to increased model size and computational overhead, and thereby limiting applicability in resource-constrained settings. For a better trade-off, we develop a task-specific feature extractor that considers both coarse- and fine-grained cues, tailored to diffusion generative traces. Benefiting from the proposed Dynamic Assembly Strategy, this advanced feature extraction can be employed without concern for memory limitations.

\begin{figure}[!t]
    \centering
    \includegraphics[width=\linewidth]{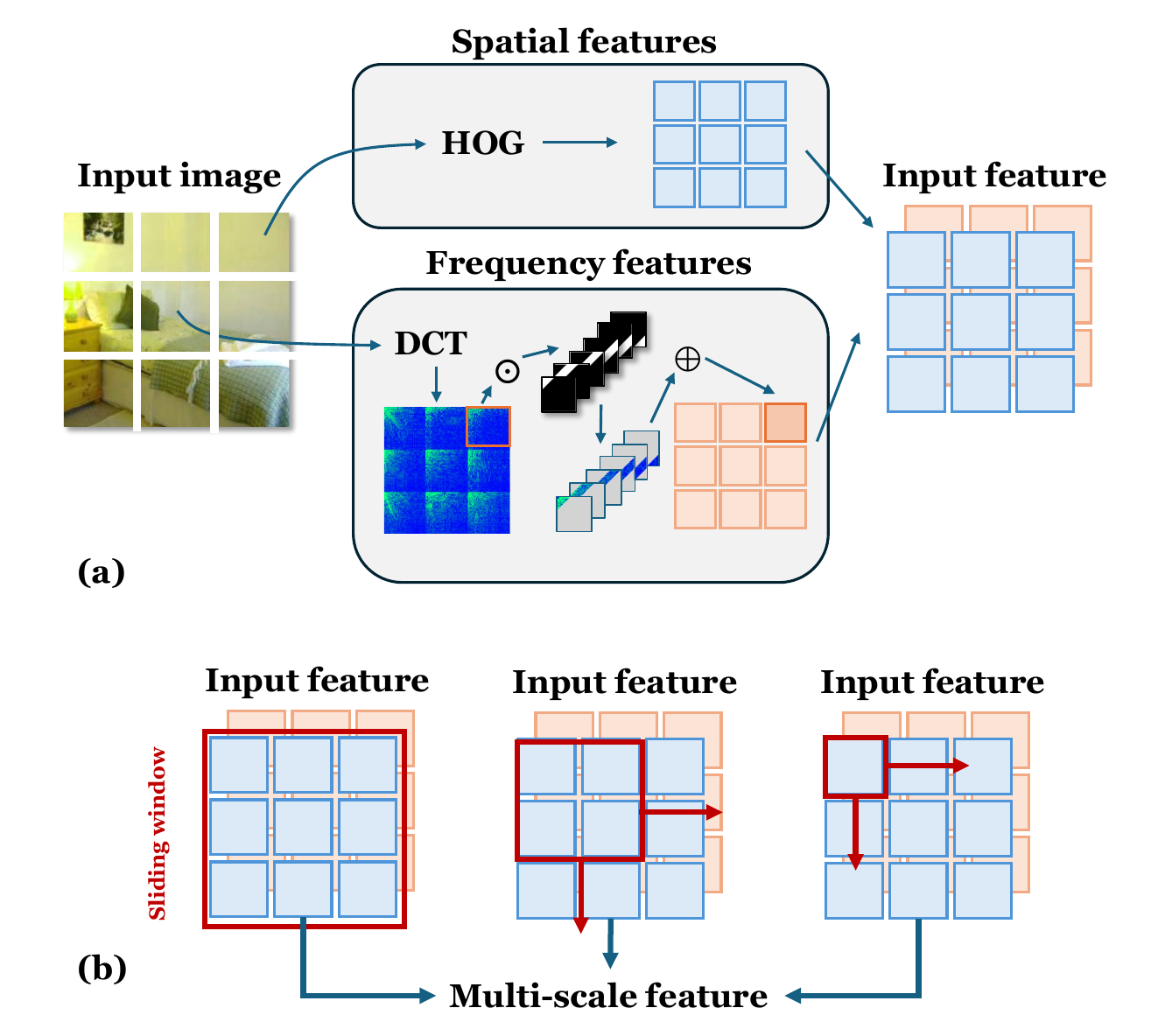}
    \vspace{-0.4cm}
    \caption{\small The detailed process of Task-specific Feature Extraction. (a) The input image is first divided into multiple patches. In each patch, we use HOG to extract spatial features. We also extract the frequency features of each patch and combine these two types of features together. (b) We employ sliding windows on batches to obtain multi-scale features.}
    \label{fig:feature}
\end{figure}

\smallskip
\noindent\textbf{Patches and Multi-scales.}
Inspired by Transformers~\cite{NIPS2017_3f5ee243}, we divide the images into $n \times n$ non-overlapping patches and extract features within each patch. Compared to ForensicsForest~\cite{lu2024forensicsforest}, our strategy focuses more on local parts, enabling the capture of fine-grained cues.  
To incorporate cues at different levels, we further introduce a multi-scale fusion scheme that integrates patch features across multiple spatial scales. 
Specifically, we construct a set of sliding windows with window sizes $\{m_i \times m_i\}^t_{i=1}$ and strides $\{s_i\}^t_{i=1}$. For each window, the features of the covered patches are averaged to form a fused representation. When $m_i=1,s_i=1$, this operation reduces to the original per-patch feature extraction, whereas $m_i=n,s_i=1$ corresponds to global feature aggregation over the entire image. By appropriately selecting $m_i$ and $s_i$, we obtain a set of features at multiple scales. Then we concatenate them as the final input feature. 
Next, we describe the feature extraction process within each patch. This process is shown in Fig.~\ref{fig:feature} (b).

For efficiency, we simply average the features within each window rather than recomputing them at every scale. This solution significantly reduces computational complexity while preserving the effectiveness of the extracted features.

\smallskip
\noindent\textbf{Patch Features.} 
For each patch, we consider extracting both spatial and frequency features. 
To extract spatial features, we utilize Histogram of Oriented Gradients (HOG)~\cite{1467360}, a classic descriptor that characterizes local gradient distributions. Generated images often exhibit subtle imperfections in regional shapes and textures, and HOG is well-suited to reveal such artifacts. Specifically, each patch is first converted to grayscale and partitioned into fixed-size cells. HOG is then computed for each cell, and the histograms of all cells are combined to form the spatial feature representation of the patch.

Frequency features have been shown to be effective in detecting generated images, as the generative artifacts are often noticeable in the frequency domain~\cite{NEURIPS2024_4f92d2f4,10004978,Doloriel2024FrequencyMF}. In line with the spatial feature extraction process, we also capture frequency features at the patch level. Unlike prior methods that rely on a global DCT transform, we adapt the Local Frequency Statistics (LFS) descriptors introduced in F$^3$-NET\cite{10.1007/978-3-030-58610-2_6}. Specifically, for each patch, we first apply the Discrete Cosine Transform (DCT) to obtain a frequency map. We then isolate different frequency components using a set of frequency filters $\{f_i\}^q_{i=1}$. We average the coefficients along the column dimension within each filtered component and concatenate the resulting vectors from all components. This process is illustrated in Fig.~\ref{fig:feature} (a). 
\section{Experiments}
\label{sec:experiment}

\begin{table*}[!t]
\centering
\setlength{\tabcolsep}{4pt}
\caption{\small Detection performance of different methods (ACC\%/AUC\%).}
\vspace{-0.2cm}
\label{table1}

\resizebox{\textwidth}{!}{
\begin{tabular}{c c
                c c c c c c 
                c c c c c c }
\toprule
\multirow{3}{*}{Eval set $\downarrow$} & 
Train set $\to$ &
\multicolumn{6}{c}{LSUN-B (\textcolor{gen}{ADM})} &
\multicolumn{6}{c}{Imagenet (\textcolor{gen}{ADM})} \\
\cmidrule(lr){2-2} \cmidrule(lr){3-8} \cmidrule(lr){9-14}
 & \multirow{2}{*}{Detector $\to$} &
\multirow{2}{*}{CNNDet} & AERO- & \multirow{2}{*}{DIRE} & Fake- & \multirow{2}{*}{FIRE} & \cellcolor{hl}\textbf{\texttt{DAF}} &  
\multirow{2}{*}{CNNDet} & AERO- & \multirow{2}{*}{DIRE} & Fake- & \multirow{2}{*}{FIRE} & \cellcolor{hl}\textbf{\texttt{DAF}} \\
& &  & BLADE &  & Inversion & & \cellcolor{hl}(Ours) &  &
BLADE &  & Inversion & & \cellcolor{hl}(Ours) \\

\midrule

\multirow{2}{*}{Imagenet} & \textcolor{gen}{ADM} & 73.4/57.6 & 99.1/98.3 & 99.4/96.4 & 100.0/99.8 & 100.0/100.0 & \cellcolor{hl}96.5/99.9 & 99.6/99.3 & 100.0/100.0 & 100.0/100.0 & 100.0/100.0 & 100.0/100.0 & \cellcolor{hl}99.5/100.0\\
         & \textcolor{gen}{SD‑v1} & 67.3/53.4 & 98.7/97.4 & 98.3/95.2 &  99.7/97.6 & 100.0/100.0 & \cellcolor{hl}94.7/98.9 & 87.2/84.7 & 99.7/98.3 & 100.0/99.6 & 100.0/100.0 & 100.0/100.0 & \cellcolor{hl}91.6/99.6 \\

\midrule

\multirow{9}{*}{LSUN‑B}   & \textcolor{gen}{ADM}   & 96.8/93.7 & 100.0/100.0 & 100.0/100.0 & 100.0/100.0 & 100.0/100.0 & \cellcolor{hl}99.9/100.0 & 76.0/54.2 & 98.4/97.5 & 100.0/99.7 & 100.0/100.0 & 100.0/100.0 & \cellcolor{hl}99.6/100.0 \\
         & \textcolor{gen}{DDPM}  & 75.3/55.1 & 98.9/97.8 & 100.0/100.0 & 100.0/100.0 & 100.0/99.8 & \cellcolor{hl}99.9/100.0 & 68.4/45.1 & 99.1/98.2 & 100.0/97.7 & 100.0/100.0 & 100.0/100.0 & \cellcolor{hl}99.7/100.0 \\
         & \textcolor{gen}{IDDPM} & 76.2/49.5 & 99.7/98.2 & 100.0/100.0 &  99.8/98.4 & 100.0/100.0 & \cellcolor{hl}100.0/100.0 & 77.6/53.8 & 99.5/98.7 & 100.0/100.0 & 98.6/97.9 & 100.0/100.0 & \cellcolor{hl}99.9/100.0 \\
         & \textcolor{gen}{PNDM}  & 81.1/49.0 & 99.2/97.9 &  99.7/88.6 & 100.0/99.7 & 100.0/100.0 & \cellcolor{hl}100.0/100.0 & 73.2/38.7 & 97.9/97.4 & 100.0/100.0 & 100.0/100.0 & 100.0/100.0 & \cellcolor{hl}99.9/100.0 \\
         & \textcolor{gen}{SD‑v2} & 78.7/50.4 & 98.3/97.5 & 100.0/100.0 & 100.0/100.0 & 100.0/100.0 & \cellcolor{hl}99.2/100.0 & 82.4/63.1 & 99.0/97.8 & 100.0/100.0 & 100.0/99.9 & 100.0/100.0 & \cellcolor{hl}98.7/100.0 \\
         & \textcolor{gen}{LDM} & 59.4/49.2 & 99.8/98.5 & 100.0/100.0 & 100.0/99.8 & 100.0/100.0 & \cellcolor{hl}99.9/100.0 & 60.8/47.5 & 98.6/97.7 & 100.0/98.6 & 100.0/100.0 & 100.0/100.0 & \cellcolor{hl}99.6/100.0 \\
         & \textcolor{gen}{VQD} & 71.5/51.9 & 98.5/97.2 & 100.0/99.8 & 100.0/100.0 & 100.0/100.0 &\cellcolor{hl}100.0/100.0 & 72.1/50.2 & 99.8/98.4 & 100.0/100.0 & 100.0/99.7 & 100.0/100.0 & \cellcolor{hl}99.9/100.0 \\
         & \textcolor{gen}{IF} & 78.0/50.3 & 99.3/97.6 & 100.0/100.0 & 100.0/99.9 & 100.0/100.0 & \cellcolor{hl}100.0/100.0 & 75.6/49.8 & 98.9/97.3 & 100.0/100.0 & 100.0/99.9 & 100.0/100.0 &\cellcolor{hl}100.0/100.0 \\
         & \textcolor{gen}{DALLE‑2} & 78.3/67.1 & 99.0/98.1 & 100.0/98.2 & 100.0/100.0 & 100.0/100.0 & \cellcolor{hl}99.7/100.0 & 77.1/57.3 & 99.2/98.5 & 100.0/99.9 & 100.0/100.0 & 100.0/100.0 & \cellcolor{hl}98.9/100.0 \\
         & \textcolor{gen}{Mid.} & 86.1/74.6 & 97.9/97.3 & 100.0/100.0 & 100.0/99.8 & 100.0/100.0 & \cellcolor{hl}100.0/100.0 & 87.9/73.2 & 98.7/97.9 & 100.0/100.0 & 100.0/99.7 & 100.0/100.0 & \cellcolor{hl}99.9/100.0 \\

\midrule
\textbf{Average} & & 76.8/58.5 & 99.0/98.0 & 99.8/98.2 & 100.0/99.6 & 100.0/100.0 & \cellcolor{hl}99.2/100.0 & 78.2/59.7 & 99.1/98.1 & 100.0/99.6 & 99.9/99.8 & 100.0/100.0 & \cellcolor{hl}98.9/100.0  \\

\bottomrule
\end{tabular}
}
\end{table*}

\subsection{Experimental Settings}
\textbf{Datasets.}
The DiffusionForensics~\cite{Wang_2023_ICCV} is a large-scale recent dataset for diffusion-generated image detection. Following~\cite{Chu_2025}, we validate our method in LSUN-B and ImageNet subsets. The training set of the LSUN-B subset contains 40,000 real images and 40,000 fake images generated by ADM~\cite{NEURIPS2021_49ad23d1}. Its testing set contains 1,000 real images and various number (100 $\sim$ 1000) of fake images generated by ADM, DDPM~\cite{NEURIPS2020_4c5bcfec}, IDDPM~\cite{pmlr-v139-nichol21a}, PNDM~\cite{liu2022pseudo}, SD-v2~\cite{Rombach_2022_CVPR}, ~LDM\cite{Rombach_2022_CVPR}, VQD~\cite{Gu_2022_CVPR}, IF~\cite{NEURIPS2022_ec795aea}, DALLE-2~\cite{ramesh2022hierarchical}, and Mid.\footnote{https://www.midjourney.com}. In the ImageNet subset, its training set also contains 40,000 real images and 40,000 fake images generated by ADM, and its testing set contains 5000 real images and 5000 fake images generated by ADM and 10000 images generated by SD-v1~\cite{Rombach_2022_CVPR}. GenImage~\cite{NEURIPS2023_f4d4a021} is a large-scale benchmark for generated image detection, consisting of images from the 1,000 ImageNet classes synthesized by eight state-of-the-art generators spanning both academic models (\eg, Stable Diffusion) and commercial systems (\eg, Midjourney). Chameleon~\cite{yan2025a} is a test set for AI-generated image detection, which specifically collects generated images that are highly similar in appearance to real images.  More details can be found in the \textit{Supplementary}.

\smallskip\noindent\textbf{Implementation Details.}
The proposed DAF is trained and tested only using an Intel Core-i7 12700 CPU, with the input size as $256 \times 256$. Specifically, each deep forest model contains $l=3$ layers, where each layer consists of $A=2$ random forests and $B = 2$ completely random forests. In the dynamic assembly stage, the number of selected forest models is set to $v=3$, and the batch size, \ie, the proportion of sampled data, is set to $p = 10\%$. The maximum number of iterations is set to $N = 10$, and the weight update factor is set to $\theta = 1.5$. Following~\cite{Chu_2025}, we employ two standard evaluation metrics: the area under the ROC curve (AUC) and classification accuracy (ACC).

\subsection{Results}

\textbf{Compared with DNN-based Methods.}
Following~\cite{Chu_2025}, our method is compared with many recent diffusion-generated image detection methods, including CNNDet~\cite{wang2019cnngenerated}, AEROBLADE~\cite{Ricker_2024_CVPR}, DIRE~\cite{Wang_2023_ICCV}, FakeInversion~\cite{Cazenavette_2024_CVPR}, and FIRE~\cite{Chu_2025}. The results are shown in Table~\ref{table1}. Note that entries highlighted in \textcolor{gen}{this color} denote the generation model, \eg, \textcolor{gen}{ADM} and \textcolor{gen}{SD-v1}. Columns 3 to 8 represent the performance of all detection methods trained on LSUN-B and tested on both datasets, while Columns 9 to 14 report results for methods trained on Imagenet and tested on both datasets. On average, our \texttt{\textbf{DAF}} achieves 99.2\% and 98.9\% ACC under two training settings, respectively, and reaches 100.0 \% AUC in both cases. These results outperform CNNDet and are comparable to those of other DNN-based methods, thereby validating the effectiveness of the proposed forest-based model.

Furthermore, we follow the experimental setup in~\cite{yan2025a}, training the model on the SDv1.4 subset of GenImage and evaluating it on the Chameleon dataset, comparing it with CNNDet~\cite{wang2019cnngenerated}, FreDect~\cite{pmlr-v119-frank20a}, Fusing~\cite{9897820}, GramNet~\cite{9157447}, LNP~\cite{10.1007/978-3-031-19781-9_6}, UniFD~\cite{Ojha_2023_CVPR}, DIRE~\cite{Wang_2023_ICCV}, PatchCraft~\cite{Zhong2023RichAP}, NPR~\cite{10658459} and AIDE~\cite{yan2025a}. This setting presents a more challenging scenario, and as shown in Table~\ref{Tab:resultsonChameleon}, the performance of most methods is generally limited, ranging between 55\% and 60\%. In contrast, \texttt{\textbf{DAF}} achieves 61.14\%, outperforming the majority of compared methods and performing on par with the current state-of-the-art. These results demonstrate that our method retains strong detection capabilities even in challenging conditions.

\begin{table}[!t]
\centering
\small
\caption{\small Detection performance of different methods on Chameleon dataset (ACC\%).}
\vspace{-0.2cm}
\label{Tab:resultsonChameleon}

\resizebox{0.8\linewidth}{!}{
\begin{tabular}{p{5cm} c}
\toprule
 Detector & Acc.(\%) \\
 \midrule
 CNNDet & 60.11\\
 FreDect & 56.86\\
 Fusing & 57.07\\
 GramNet & 60.95\\
 LNP & 55.63\\
 UniFD & 55.62\\
 DIRE & 59.71\\
 PatchCraft & 56.32\\
 NPR & 58.13\\
 AIDE & \textbf{62.60}\\
 \cellcolor{hl}\textbf{\texttt{DAF}}(Ours) & \cellcolor{hl}61.14\\

\bottomrule
\end{tabular}
}
\end{table}

\begin{table}[!t]
\centering
\caption{\small Comparison results of our method and ForensicsForest on a fair set (ACC\%/AUC\%).}
\vspace{-0.2cm}
\label{comparewithforensicsforest}

\setlength{\tabcolsep}{4pt} 
\small                      

\resizebox{0.95\columnwidth}{!}{%
\begin{tabular}{c c c c c c}
\toprule
\multirow{3}{*}{Eval set $\downarrow$} & 
Train set $\to$ &
\multicolumn{2}{c}{LSUN-B (\textcolor{gen}{ADM})} &
\multicolumn{2}{c}{Imagenet (\textcolor{gen}{ADM})} \\
\cmidrule(lr){2-2} \cmidrule(lr){3-4} \cmidrule(lr){5-6}
 & \multirow{2}{*}{Detector  $\to$} &
Forensics & \textbf{\texttt{DAF}} & Forensics & \textbf{\texttt{DAF}}\\
& & Forest & (Ours) & Forest & (Ours) \\

\midrule

\multirow{2}{*}{Imagenet} & \textcolor{gen}{ADM} & 87.5/98.2 & \cellcolor{hl}90.2/97.9 & 97.3/99.7 & \cellcolor{hl}97.8/99.9 \\
         & \textcolor{gen}{SD‑v1} & 84.5/91.6 & \cellcolor{hl}89.6/94.0 & 86.3/96.2 & \cellcolor{hl}79.7/98.3 \\

\midrule

\multirow{9}{*}{LSUN‑B} & \textcolor{gen}{ADM} & 92.0/99.0 & \cellcolor{hl}98.0/100.0 & 94.0/98.1 & \cellcolor{hl}97.5/99.9 \\
         & \textcolor{gen}{DDPM}  & 91.0/98.6 & \cellcolor{hl}98.9/99.9 & 94.4/98.3 & \cellcolor{hl}97.2/99.9 \\
         & \textcolor{gen}{IDDPM} & 92.5/99.7 & \cellcolor{hl}98.5/100.0 & 95.5/99.4 & \cellcolor{hl}99.5/100.0 \\
         & \textcolor{gen}{PNDM}  & 91.5/99.2 & \cellcolor{hl}99.5/100.0 & 91.5/97.3 & \cellcolor{hl}98.5/100.0 \\
         & \textcolor{gen}{SD‑v2} & 91.0/97.9 & \cellcolor{hl}92.5/99.7 & 88.0/96.3 & \cellcolor{hl}84.5/99.4 \\
         & \textcolor{gen}{LDM} & 91.5/99.0 & \cellcolor{hl}99.5/100.0 & 96.0/99.7 & \cellcolor{hl}99.0/100.0 \\
         & \textcolor{gen}{VQD} & 91.5/99.0 & \cellcolor{hl}98.0/100.0 & 91.5/97.6 & \cellcolor{hl}95.0/99.8 \\
         & \textcolor{gen}{IF} & 92.0/99.4 & \cellcolor{hl}99.5/100.0 & 97.0/99.4 & \cellcolor{hl}99.5/100.0 \\
         & \textcolor{gen}{DALLE‑2} & 88.7/96.1 & \cellcolor{hl}94.0/99.7 & 86.0/93.9 & \cellcolor{hl}88.7/99.4 \\
         & \textcolor{gen}{Mid.} & 86.4/98.3 & \cellcolor{hl}97.3/99.8 & 95.5/98.0 & \cellcolor{hl}96.4/99.4 \\

\midrule
\textbf{Average} & & 90.0/98.0 & \cellcolor{hl}96.3/99.3 & 92.8/97.8 & \cellcolor{hl}94.4/99.7 \\
\bottomrule
\end{tabular}}
\end{table}

\begin{table*}[!t]
\centering
\caption{\small Effect of multi-scale features.}
\vspace{-0.2cm}
\label{tab:differentscale}

\small                      

\resizebox{0.8\textwidth}{!}{%
\begin{tabular}{c c c c c c c c}
\toprule
\multirow{3}{*}{Eval set $\downarrow$} & 
Train set $\to$ &
\multicolumn{3}{c}{LSUN-B (\textcolor{gen}{ADM})} &
\multicolumn{3}{c}{Imagenet (\textcolor{gen}{ADM})} \\
\cmidrule(lr){2-2} \cmidrule(lr){3-5} \cmidrule(lr){6-8}
 & Detector $\to$ &
Single & Multi (Recalc.) & Multi (Avg.) & Single & Multi (Recalc.) & Multi (Avg.) \\

\midrule

\multirow{2}{*}{Imagenet} & \textcolor{gen}{ADM} & 96.3/99.7 & 94.7/99.9 & 96.5/99.9 & 98.1/100.0 & 99.7/100.0 & 99.5/100.0\\
         & \textcolor{gen}{SD‑v1} & 93.7/98.6 & 94.3/98.6 & 94.7/98.9 & 90.1/98.5 & 84.7/99.3 & 91.6/99.6 \\

\midrule

\multirow{9}{*}{LSUN‑B} & \textcolor{gen}{ADM} & 99.8/99.9 & 99.9/100.0 & 99.9/100.0 & 99.5/99.9 & 98.3/100.0 & 99.6/100.0 \\
         & \textcolor{gen}{DDPM}  & 99.7/99.9 & 99.8/100.0 & 99.9/100.0 & 99.5/99.9 & 98.9/100.0 & 99.7/100.0 \\
         & \textcolor{gen}{IDDPM} & 99.9/100.0 & 99.9/100.0 & 100.0/100.0 & 99.7/100.0 & 99.1/100.0 & 99.9/100.0 \\
         & \textcolor{gen}{PNDM}  & 99.8/100.0 & 99.9/100.0 & 100.0/100.0 & 99.6/100.0 & 99.7/100.0 & 99.9/100.0 \\
         & \textcolor{gen}{SD‑v2} & 99.0/100.0 & 99.0/100.0 & 99.2/100.0 & 99.2/100.0 & 89.0/100.0 & 98.7/100.0 \\
         & \textcolor{gen}{LDM} & 99.8/100.0 & 99.9/100.0 & 99.9/100.0 & 99.0/99.9 & 99.7/100.0 & 99.6/100.0 \\
         & \textcolor{gen}{VQD} & 99.8/100.0 & 99.9/100.0 & 100.0/100.0 & 99.6/100.0 & 99.2/100.0 & 99.9/100.0 \\
         & \textcolor{gen}{IF} & 99.8/100.0 & 99.9/100.0 & 100.0/100.0 & 99.6/100.0 & 100.0/100.0 & 100.0/100.0 \\
         & \textcolor{gen}{DALLE‑2} & 99.3/100.0 & 99.5/100.0 & 99.7/100.0 & 99.5/100.0 & 99.5/100.0 & 98.9/100.0 \\
         & \textcolor{gen}{Mid.} & 99.7/100.0 & 99.8/100.0 & 100.0/100.0 & 99.5/100.0 & 98.0/99.9 & 99.9/100.0 \\

\midrule
\textbf{Average} & & 98.9/99.8 & 98.9/99.9 & \textbf{99.2/100.0} & 98.6/99.9 & 97.2/99.9 & \textbf{98.9/100.0}  \\
\bottomrule
\end{tabular}
}
\end{table*}

\smallskip\noindent\textbf{Compared with ForensicsForest.}
ForensicsForest~\cite{lu2024forensicsforest} is another forest-based method for detecting generated images. However, due to the large scale of the datasets used in our experiments, ForensicsForest is impractical to train on the full dataset, primarily because of the unresolved data-loading bottleneck discussed in Sec.~\ref{sec:intro}.  
To enable a fair comparison, we reduce the task difficulty by randomly sampling 10\% of the training and test sets to form a smaller dataset. With the reduced data scale, we disable the dynamic assembly strategy and instead build deep forest models using all available training samples. Under this setting, the comparison primarily reflects the effectiveness of the feature extraction strategies. As shown in Table~\ref{comparewithforensicsforest}, our method notably outperforms ForensicsForest, achieving an improvement of 6.3\% in ACC and 1.3\% in AUC on the LSUN-B dataset, and achieving a 1.6\% increase in ACC and a 1.9\% increase in AUC on the Imagent dataset.

\subsection{Ablation studies}
\textbf{Effect of Multi-scale Features.}
To verify the effectiveness of the proposed multi-scale feature extraction, we compared its performance with a single scale. The specific experimental setup is as follows: 1) For the single-scale case, we only process the images on the {$16 \times 16$} patch division. 2) For the multi-scale case, we design several sliding windows with sizes of {$8 \times 8$, $4 \times 4$, $2 \times 2$, and $1 \times 1$} patches, and apply them on the {$16 \times 16$} patch division without overlapping. Then we average pool the features of patches inside sliding windows. 3) In addition, we explore another multi-scale strategy by recalculating the features inside each sliding window. The experimental results in Table~\ref{tab:differentscale} show that using multi-scale feature extraction is superior to using only one scale, and simply averaging the features inside sliding windows performs slightly better. Thus, we adopt this strategy in the main experiments.

\begin{figure}[h]
  \centering
  \includegraphics[width=0.7\linewidth]{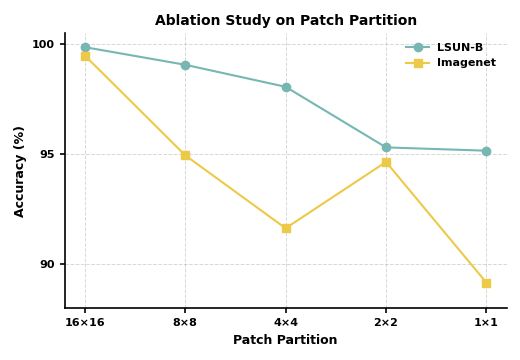}
  \vspace{-0.5cm}
  \caption{\small Effect of the number of image patch partitions.}
  \label{patch_partition}
\end{figure}

We also investigate the effect of patch size on performance, considering patch numbers of 16×16, 8×8, 4×4, 2×2 and 1×1. The test results, presented in Fig.~\ref{patch_partition}, show that detection accuracy decreases as the number of patches decreases. The best performance is achieved with 16×16 patches, which is therefore adopted in our main experiments.

\begin{figure}[!t]
  \centering
  \includegraphics[width=\linewidth]{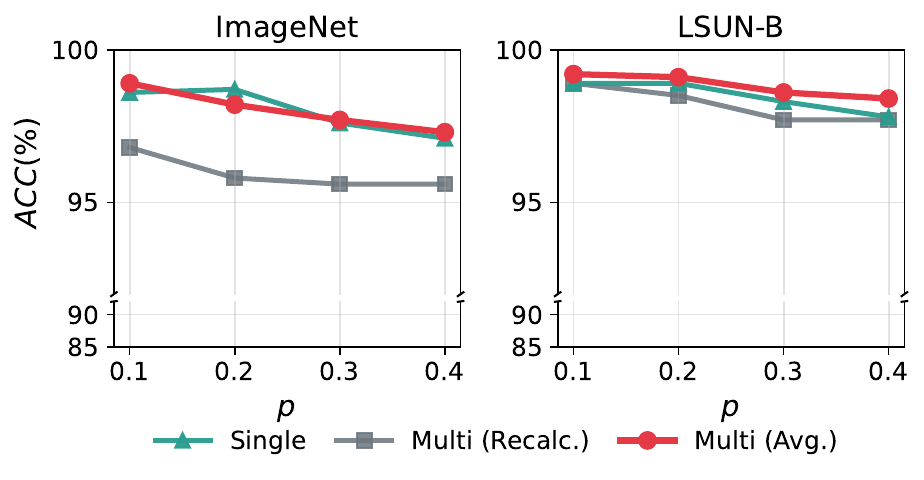}
  \vspace{-0.8cm}
  \caption{\small The effect of different $p$ values on ACC (\%) on ImageNet and LSUN-B datasets with three settings: Single, Multi (Recalc.), Multi (Avg.).}
  \label{p_acc_imagenet_lsunb}
\end{figure}

\smallskip\noindent\textbf{Proportion of Sampled Data $p$.}
The proportion of sampled data $p$ in our method plays a role analogous to the ``batch size''. Specifically, $p = 1$ means loading all data into memory, while $p=0$ denotes no sampling. A larger $p$ allows the model to access more data simultaneously but incurs higher memory consumption, while a smaller $p$ reduces memory requirements at the cost of potentially degraded effectiveness. In our experiments, we vary 
$p$ within the range $[0.1, 0.4]$. Fig.~\ref{p_acc_imagenet_lsunb} illustrates the average ACC changes of \texttt{\textbf{DAF}} on the test set. {The detailed experimental results of the three experimental setups under different $p$-values can be found in \textit{Supplementary}.}
It can be seen that with $p$ increasing, the performance slightly drops (no more than $1\%$).  
This is because, in our method, a smaller $p$ leads to greater diversity among forest models constructed across different batches, analogous to introducing multiple complementary experts. Ensembling these diverse forest models, therefore, leads to improved performance. Based on these observations, we set $p = 0.1$ in the main experiments.

\subsection{Further Analysis and Discussion}
\begin{figure}[!t]
  \centering
  \includegraphics[width=\linewidth]{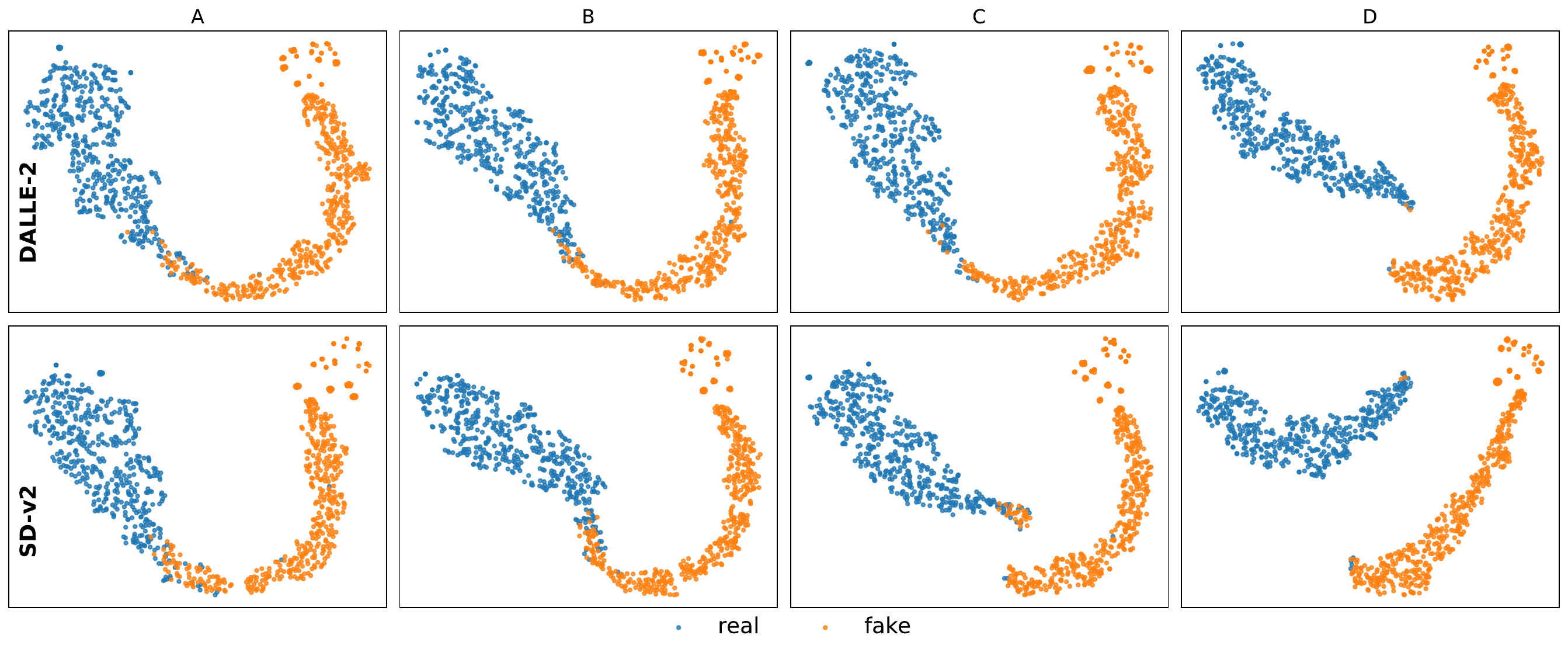}
  \vspace{-0.8cm}
  \caption{\small The visualization results trained on the LSUN-B dataset and tested on DALLE-2 and SD-v2 of the LSUN-B dataset.}
  \label{TSNE_LSUNB}
  \vspace{-0.3cm}
\end{figure}

\begin{figure}[!t]
  \centering
  \includegraphics[width=\linewidth]{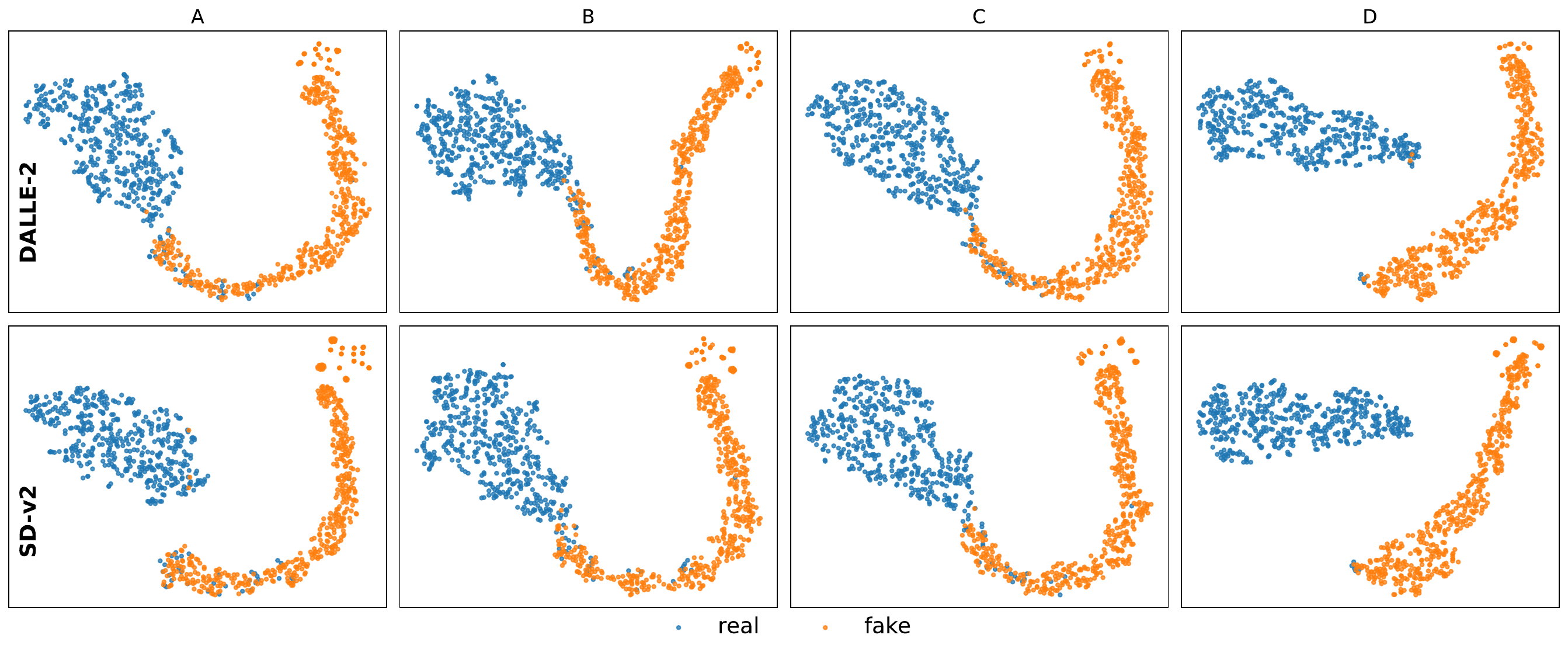}
  \vspace{-0.8cm}
  \caption{\small The visualization results trained on the Imagenet dataset and tested on DALLE-2 and SD-v2 of the LSUN-B dataset.}
  \label{TSNE_Imagenet}
\end{figure}

\textbf{Feature Distribution Visualization.} 
In this part, we visualize the feature distributions to illustrate the performance gains brought by the proposed dynamic assembly strategy for deep forest models.
Specifically, we randomly sample 500 real and 500 fake images from the DALLE-2 and SD-v2 test sets of LSUN-B. We consider two settings: features extracted from forest models constructed within individual batches before assembly, and features obtained after dynamic assembly. For visualization, we use the features from the last layer of each individual forest model. Fig.~\ref{TSNE_LSUNB} and Fig.~\ref{TSNE_Imagenet} illustrate the results. The left three columns (A, B, C) correspond to forest models before assembly, while the rightmost column (D) shows the final assembled model. As can be observed, dynamic assembly leads to a visibly larger separation between real and fake samples, demonstrating the effectiveness of the proposed strategy.

\begin{figure*}[!t]
  \centering
  \includegraphics[width=0.95\textwidth]{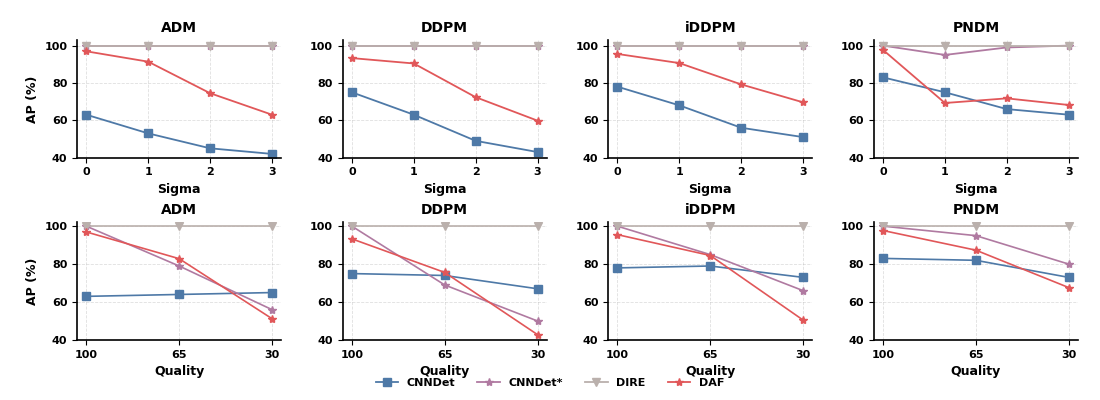}
  \vspace{-0.4cm}
  \caption{\small Robustness under various perturbations, including Gaussian blur (top) and JPEG compression (bottom). }
  \label{robustness}
\end{figure*}

\smallskip\noindent\textbf{Robustness.} 
Following the experimental settings in DIRE~\cite{Wang_2023_ICCV}, we evaluate the robustness of various methods under two types of degradation: Gaussian blur and JPEG compression. Gaussian blur is applied at three intensities ($\delta$ = 1, 2, 3), while JPEG compression is tested at two quality levels (quality = 65, 30). In this study, we compare CNNDet~\cite{wang2019cnngenerated}(* denotes reproduced training on the LSUN-Bedroom-ADM subset of DiffusionForensics), DIRE, and our method, \texttt{\textbf{DAF}}. As shown in Fig.~\ref{robustness}, the performance of DAF gradually decreases under both Gaussian blur and JPEG compression, but it remains relatively stable and competitive compared to the DNN-based method CNNDet. Nevertheless, due to inherent differences between traditional machine learning and deep learning, DAF does not surpass more recent advanced methods like DIRE when severe degradations are introduced.

\begin{table}[!t]
\centering
\small
\caption{\small Discussion: ACC (\%) performance of our method with($w.$) or without($wo.$) on-the-fly data augmentation.}
\vspace{-0.2cm}
\label{table5}

\resizebox{0.95\linewidth}{!}{
\begin{tabular}{c c
                c c 
                c c}
\toprule
\multirow{3}{*}{Eval set $\downarrow$} & 
Train set $\to$ &
\multicolumn{2}{c}{LSUN-B (\textcolor{gen}{ADM})} &
\multicolumn{2}{c}{Imagenet (\textcolor{gen}{ADM})} \\
\cmidrule(lr){2-2} \cmidrule(lr){3-4} \cmidrule(lr){5-6}
 & Detector $\to$ & $w.$ & $wo.$ & $w.$ & $wo.$ \\

\midrule

\multirow{2}{*}{Imagenet} & \textcolor{gen}{ADM} & 94.2 & 96.5 & 99.5 & 99.5 \\
         & \textcolor{gen}{SD‑v1} & 94.1 & 94.7 & 90.9 & 91.6 \\

\midrule

\multirow{9}{*}{LSUN‑B} & \textcolor{gen}{ADM} & 99.8 & 99.9 & 99.5 & 99.6 \\
         & \textcolor{gen}{DDPM}  & 99.7 & 99.9 & 99.9 & 99.7 \\
         & \textcolor{gen}{IDDPM} & 99.8 & 100.0 & 100.0 & 99.9 \\
         & \textcolor{gen}{PNDM}  & 99.8 & 100.0 & 99.8 & 99.9 \\
         & \textcolor{gen}{SD‑v2} & 99.6 & 99.2 & 98.3 & 98.7 \\
         & \textcolor{gen}{LDM} & 99.8 & 99.9 & 100.0 & 99.6 \\
         & \textcolor{gen}{VQD} & 99.8 & 100.0 & 99.7 & 99.9 \\
         & \textcolor{gen}{IF} & 99.8 & 100.0 & 100.0 & 100.0 \\
         & \textcolor{gen}{DALLE‑2} & 99.5 & 99.7 & 99.0 & 98.9 \\
         & \textcolor{gen}{Mid.} & 99.6 & 100.0 & 100.0 & 99.9 \\

\midrule
\textbf{Average} & & 98.8 & \textbf{99.2} & 98.9 & \textbf{98.9} \\

\bottomrule
\end{tabular}
}
\end{table}

\smallskip\noindent\textbf{Could our method use on-the-fly data augmentation?}
Traditional forest-based models require loading the entire dataset into memory, making data augmentation impractical, as it would substantially increase memory consumption. In contrast, our method adopts a batch-wise data loading and model construction paradigm, which, in theory, enables on-the-fly data augmentation, similar to DNNs. In our preliminary study, we consider three augmentation strategies: horizontal flipping, random cropping, and additive Gaussian noise. For each batch $b_i$, these augmentations are applied randomly to the samples with a probability of 0.5.
Table~\ref{table5} shows the results. Contrary to our expectations, data augmentation does not yield a notable performance improvement and even leads to a slight degradation. We attribute this behavior to the characteristics of the evaluated dataset~\cite{Chu_2025}, which has already undergone certain preprocessing and augmentation operations. Applying additional augmentations may further distort the image content, causing it to deviate from the underlying data distribution and thereby hindering the learning of subtle synthesis traces. In future work, we plan to further investigate the effect of on-the-fly data augmentation using other large-scale yet minimally processed datasets.

\smallskip\noindent\textbf{Tentative try of Integrating DNNs.}
To enhance feature extraction, we explore integrating deep feature representations into our model, using ResNet18 as the base network. Specifically, we train ResNet18 on the corresponding datasets and fuse its intermediate layer features with those extracted by our method, feeding the combined features into \texttt{\textbf{DAF}}. The results in Table~\ref{featurewithresnet} show that this approach improves performance within the same data domain. However, it substantially increases overfitting, leading to a notable decline in generalization across datasets. These findings indicate that our current model is not well-suited for direct integration with deep feature representations, and we leave the exploration of this part as future work.

\begin{table}[!t]
\centering
\small
\caption{\small Discussion: ACC (\%) performance of our method with($w.$) or without($wo.$) ResNet18 feature extraction.}
\vspace{-0.2cm}
\label{featurewithresnet}

\resizebox{0.95\linewidth}{!}{
\begin{tabular}{c c
                c c 
                c c}
\toprule
\multirow{3}{*}{Eval set $\downarrow$} & 
Train set $\to$ &
\multicolumn{2}{c}{LSUN-B (\textcolor{gen}{ADM})} &
\multicolumn{2}{c}{Imagenet (\textcolor{gen}{ADM})} \\
\cmidrule(lr){2-2} \cmidrule(lr){3-4} \cmidrule(lr){5-6}
 & Detector $\to$ & $w.$ & $wo.$ & $w.$ & $wo.$ \\

\midrule

\multirow{2}{*}{Imagenet} & \textcolor{gen}{ADM} & 54.7 & 96.5 & 99.9 & 99.5 \\
         & \textcolor{gen}{SD‑v1} & 40.3 & 94.7 & 98.3 & 91.6 \\

\midrule

\multirow{9}{*}{LSUN‑B} & \textcolor{gen}{ADM} & 100.0 & 99.9 & 50.0 & 99.6 \\
         & \textcolor{gen}{DDPM}  & 99.9 & 99.9 & 56.6 & 99.7 \\
         & \textcolor{gen}{IDDPM} & 100.0 & 100.0 & 50.0 & 99.9 \\
         & \textcolor{gen}{PNDM}  & 99.8 & 100.0 & 50.0 & 99.9 \\
         & \textcolor{gen}{SD‑v2} & 100.0 & 99.2 & 50.0 & 98.7 \\
         & \textcolor{gen}{LDM} & 100.0 & 99.9 & 50.0 & 99.6 \\
         & \textcolor{gen}{VQD} & 100.0 & 100.0 & 50.0 & 99.9 \\
         & \textcolor{gen}{IF} & 100.0 & 100.0 & 50.0 & 100.0 \\
         & \textcolor{gen}{DALLE‑2} & 99.9 & 99.7 & 66.7 & 98.9 \\
         & \textcolor{gen}{Mid.} & 100.0 & 100.0 & 90.9 & 99.9 \\

\bottomrule
\end{tabular}
}
\vspace{-0.3cm}
\end{table}

\smallskip\noindent\textbf{Pros and Cons.}
Compared to mainstream DNN-based methods, our method inherits several merits of traditional machine learning models, including simplicity, lightweight design, and cost efficiency. Meanwhile, as our method is built upon decision models, few parameters are introduced, which endows it with a natural robustness against adversarial attacks~\cite{goodfellow2014explaining}, which typically exploit the large parameter space of deep neural networks. Consequently, our method can be considered more secure than DNN-based counterparts.

On the other hand, since our method is fully deployed on CPUs, the feature extraction stage may incur higher runtime compared with GPU-based implementations. Therefore, we would like to extend the feature extraction pipeline to GPUs and develop a hybrid CPU–GPU framework to further improve computational efficiency in future work.
\section{Conclusion}
\label{sec:conclusion}

In this paper, we revisit traditional machine learning and propose a novel \texttt{\textbf{D}}ynamic \texttt{\textbf{A}}ssembly \texttt{\textbf{F}}orest model (\texttt{\textbf{DAF}}) to detect diffusion-generated images. Our method introduces two key improvements: dynamic forest assembly strategy and task-specific feature extraction, which enable batch-wise training in a manner analogous to DNNs and make it an effective diffusion-generated image detector. 
Our method is lightweight and can be deployed without GPUs, and the experimental results demonstrate its effectiveness compared to DNN-based counterparts.
{
    \small
    \bibliographystyle{ieeenat_fullname}
    \bibliography{main}
}


\end{document}